\documentclass[conference]{IEEEtran}
\IEEEoverridecommandlockouts
\usepackage{cite}
\usepackage{amsmath,amssymb,amsfonts}
\usepackage{algorithmic}
\usepackage{graphicx}
\usepackage{textcomp}
\usepackage{xcolor}
\usepackage{float}
\usepackage{booktabs}
\usepackage{multirow}
\usepackage{wrapfig}
\usepackage{tabularx}
\usepackage{array}
\usepackage{ragged2e}
\usepackage[colorlinks=true,
            linkcolor=blue,
            citecolor=blue,
            urlcolor=blue]{hyperref}
\usepackage{tikz}

\def\BibTeX{{\rm B\kern-.05em{\sc i\kern-.025em b}\kern-.08em
    T\kern-.1667em\lower.7ex\hbox{E}\kern-.125emX}}

\usepackage{titlesec}

\titlespacing*{\section}{0pt}{3pt}{1pt}
\titlespacing*{\subsection}{0pt}{2pt}{1pt}
\titlespacing*{\subsubsection}{0pt}{1pt}{0.5pt}

\begin{document}
\title{Multimodal Skin Lesion Classification with Swin Transformer and Clinical Metadata Fusion}




\author{
\IEEEauthorblockN{
Nethmi Pathirana\textsuperscript{1},
Isuru Munasinghe\textsuperscript{2},
Dileeka Alwis\textsuperscript{1}
}
\IEEEauthorblockA{
\textsuperscript{1}School of Computing, Informatics Institute of Technology, Colombo, Sri Lanka\\
\textsuperscript{2}Department of Electronic and Telecommunication Engineering, University of Moratuwa, Katubedda, Sri Lanka\\
}
}

\maketitle

\begin{abstract}
Skin lesion classification plays an important role in supporting the early diagnosis of skin cancer. However, automated analysis remains challenging due to class imbalance, inter-class similarity, and intra-class variability in dermoscopic images. This paper proposes a multimodal classification framework that combines Swin Transformer-based image features with structured clinical metadata to improve diagnostic performance through integrated visual-context learning. Experiments on a publicly available dataset show that the proposed model achieves a test accuracy of 92.55\% and a macro F1-score of 91.33\%, with strong performance across minority classes. Temperature scaling is applied as a post-hoc calibration method, resulting in a reduction in expected calibration error and improving prediction reliability, while uncertainty estimation is incorporated to further assess the confidence of model predictions. Qualitative explainability analysis further shows that the model focuses on lesion regions during inference. Therefore, the results demonstrate that multimodal fusion, combined with calibration and interpretability analysis, provides an effective and trustworthy approach for automated skin lesion classification.
\end{abstract}


\section{Introduction}

Skin cancer remains a major healthcare concern, and early identification of malignant lesions is essential for improving patient outcomes~\cite{ahmed2020malignant}. Dermoscopic imaging helps clinical diagnosis by improving the visibility of lesion structures, but visual interpretation remains challenging due to high inter-class similarity, intra-class variability, and patient-specific differences in pigmentation, anatomical site, and lesion appearance. Clinical assessment is also influenced by observer experience, which motivates computer-aided diagnostic systems that provide consistent and repeatable analysis.

Deep learning has advanced automated skin lesion classification through transfer learning and pretrained convolutional neural network (CNN) feature extraction~\cite{bassel2022automatic}. However, clinical applicability remains limited by class imbalance, poor minority-class sensitivity, dataset-specific visual biases, limited external validation, and inconsistent reporting of per-class performance. These limitations are especially important in medical classification tasks, where misclassification of rare malignant categories can have significant clinical consequences. Clinical metadata such as age, sex, and anatomical site provide complementary diagnostic context that can improve discrimination among visually similar lesions~\cite{wen2022characteristics}. Recent studies have shown that transformer-based models achieve strong performance in large-scale visual recognition tasks by learning discriminative representations from extensive image datasets~\cite{rodrigo2026benchmarking, rodrigo2025vision}. Motivated by these advances, this study investigates multimodal skin lesion classification using dermoscopic images and clinical metadata, with a focus on leakage-aware evaluation, imbalance mitigation, and clinical performance analysis. The main contributions of this work are summarized as follows.
\begin{enumerate}

    \item We develop a multimodal lesion classification framework that integrates Swin Transformer attention-based image representations with structured clinical metadata for collaborative visual-context learning.
    
    \item We integrate reliability-oriented modeling by combining calibrated probability estimation and predictive uncertainty analysis to improve confidence in automated lesion diagnosis.
    
    \item We adapt explainable artificial intelligence (XAI) techniques to the multimodal transformer architecture to provide clinically interpretable visual evidence for model predictions.
\end{enumerate}

The remainder of the paper is organized as follows: Section~\ref{sec2} presents the related work, Section~\ref{sec3} describes the proposed methodology, Section~\ref{sec4} reports the experimental results and evaluation. Finally, the paper concludes with a summary of findings and future research directions.

\section{Related Work}
\label{sec2}

\begin{figure*}[!t]
    \centering
    \includegraphics[width=\linewidth]{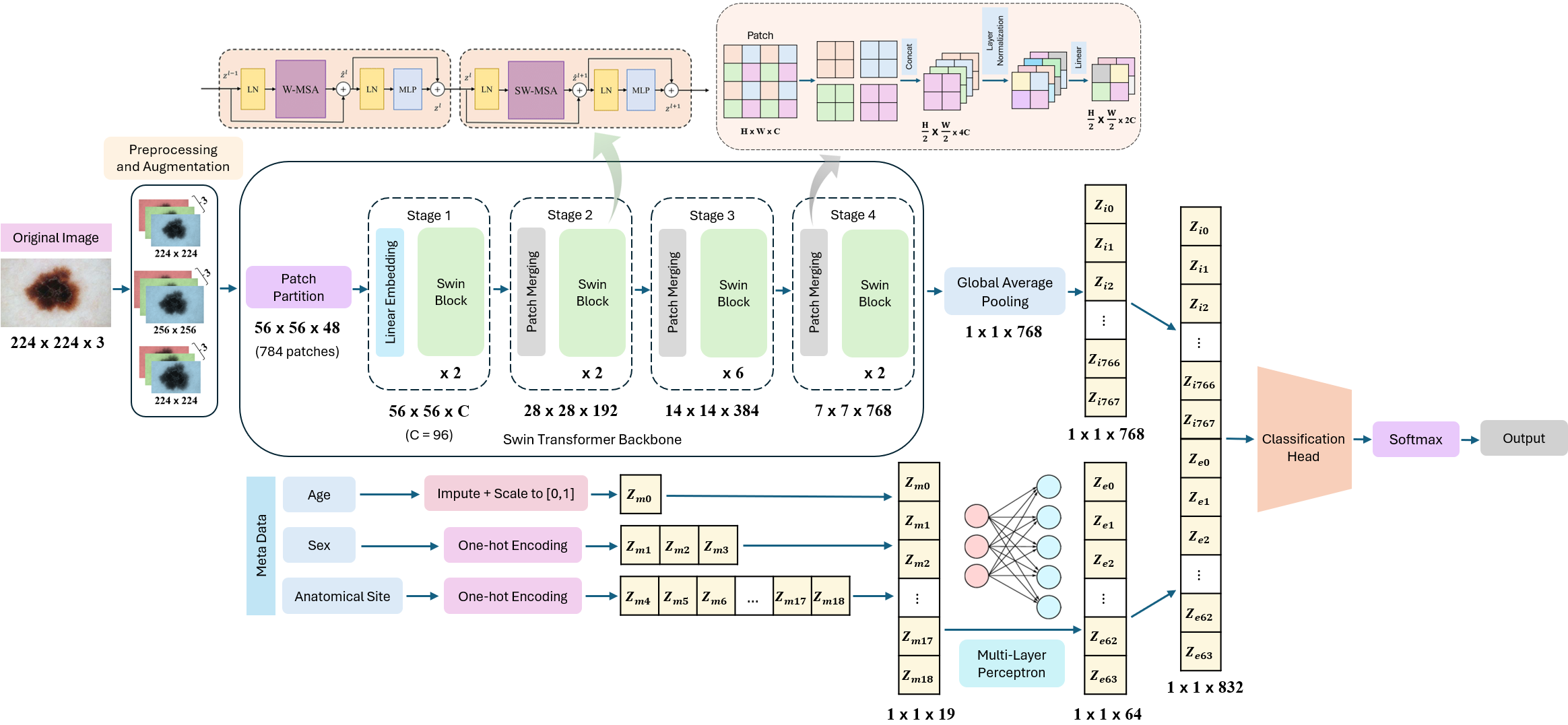}
    \caption{Proposed multimodal fusion architecture with a multi-stage Swin transformer encoder.}
    \label{multimodal_arch}
\end{figure*}

Transfer learning with pretrained convolutional networks remains the most frequently reported strategy for skin lesion classification, where performance improvements are often framed around backbone selection and fine-tuning details. Skin disease detection using deep learning~\cite{inthiyaz2023skin} illustrates this pattern through a straightforward transfer learning classifier, while early detection of skin cancer using deep learning architectures, ResNet 101 and Inception v3~\cite{demir2019early}, and deep learning based transfer learning for classification of skin cancer~\cite{bechelli2022machine} similarly evaluate popular backbones under comparable training setups. Larger benchmarking studies broaden comparisons across many architectures and demonstrate that performance can vary substantially with the chosen feature extractor and training setup~\cite{gajera2023comprehensive}. Practical deployment constraints are addressed in lightweight designs such as Squeeze MNet for low-computing Internet of Things (IoT) devices~\cite{shinde2022squeeze} and MobileNetV2-based transfer learning pipelines~\cite{rashid2022skin}. Reported gains are often limited by class imbalance that inflates overall accuracy while degrading minority class sensitivity, and many studies rely on single split evaluation or limited external testing, which weakens claims about generalization across devices.

Pipeline-based approaches extend beyond end-to-end classification by adding preprocessing, feature engineering, or multi-stage decision support. Wavelet transform-based deep residual neural network and ReLU-based extreme learning machine for skin lesion classification focus on artifact reduction and activation behavior to stabilize feature learning~\cite{alenezi2023wavelet}. Multi-stage melanoma recognition with deep residual features and hyperparameter optimization integrates deep features with feature selection and classical classifiers, which can improve discrimination in constrained settings but increases dependence on tuned components~\cite{alenezi2023multi}.
Hybrid learning that combines deep features with multiple traditional classifiers and stacked validation has also been explored to improve performance consistency~\cite{bassel2022automatic}. Joint segmentation and classification frameworks that incorporate lesion localization, including InSiNet and segmentation classification integration methods, aim to reduce background bias and force attention toward the lesion region~\cite{reis2022insinet}. These designs often add several processing steps that can introduce hidden data leakage through preprocessing choices, and they can reduce reproducibility when segmentation, enhancement, and augmentation parameters are not controlled consistently across folds.

Trust-oriented research emphasizes interpretability and calibrated decision-making beyond accuracy alone. An explainable stacked ensemble of deep learning models introduces an ensemble with explainability to support clinical reading of predictions~\cite{shorfuzzaman2022explainable}. Domain knowledge driven learning has been studied by detecting specific dermatological signs in the lesion and using those detected cues as additional evidence for classification, which helps align the decision process with clinically recognized criteria beyond general texture patterns learned from images~\cite{bian2022skin}. Limitations remain in calibration under dataset shift and in the limited use of patient context, since uncertainty and explanations are commonly derived from image signals alone, which can misrepresent ambiguity when non-visual factors would change the clinical interpretation.

\section{Methodology}
\label{sec3}

In this section, we present the proposed multimodal learning framework, including the dataset preparation strategy, architectural design, and training settings for skin lesion classification.

\subsection{Problem Formulation}
Skin lesion diagnosis in this study is posed as a supervised multimodal multi-class classification problem. Each observation is defined as a tuple $x_i=(I_i, m_i, y_i)$, where $I_i \in \mathbb{R}^{H\times W\times 3}$ denotes a dermoscopic image, $m_i \in \mathbb{R}^{d}$ is a structured clinical feature vector derived from patient age, sex, and anatomical site, and $y_i \in \{1,\ldots,C\}$ is the ground-truth diagnostic label among $C$ lesion categories. The learning objective is to estimate a parameterized mapping $f_{\theta}: (I_i, m_i) \mapsto \hat{p}_i$, where $\hat{p}_i \in [0,1]^C$ represents the predicted class probability distribution. Model parameters $\theta$ are optimized using a class-weighted cross-entropy loss to reduce bias toward majority classes in the highly imbalanced class distribution. To obtain an unbiased estimate of generalization, the data are partitioned using group-aware splitting such that samples associated with the same patient or lesion identifier do not appear in both training and test sets, limiting identity leakage.

\subsection{Dataset Description}

The experiments are conducted using the Human Against Machine with 10000 training images (HAM10000) dataset~\cite{tschandl2018ham10000}, a publicly available dermoscopic skin lesion classification benchmark containing 10,015 images categorized into seven diagnostic classes: melanocytic nevi (nv), melanoma (mel), benign keratosis-like lesions (bkl), basal cell carcinoma (bcc), actinic keratoses (akiec), vascular lesions (vasc), and dermatofibroma (df). In addition to image data, the dataset includes clinical metadata such as patient age, sex, and lesion anatomical site, which provide complementary contextual information for multimodal classification. However, HAM10000 shows considerable class imbalance, with nv forming the majority class and rare categories such as df and vasc having relatively few samples, as summarized in Table~\ref{ham10000_class_dist}.

\begin{table}[t]
\centering
\caption{Class distribution of the HAM10000 dataset.}
\label{ham10000_class_dist}
\begin{tabular}{cccrr}
\hline
\textbf{ID} & \textbf{Class (dx)} & \textbf{Clinical Group} & \textbf{Count} & \textbf{Percentage (\%)} \\
\hline
0 & akiec & Pre-malignant &  327 &  3.27 \\
1 & bcc   & Malignant     &  514 &  5.13 \\
2 & bkl   & Benign        & 1099 & 10.97 \\
3 & df    & Benign        &  115 &  1.15 \\
4 & mel   & Malignant     & 1113 & 11.11 \\
5 & nv    & Benign        & 6705 & 66.95 \\
6 & vasc  & Benign        &  142 &  1.42 \\
\hline
\end{tabular}
\end{table}

\subsection{Leakage-Aware Data Partitioning}

Data leakage was mitigated using a group-based splitting strategy, as dermoscopic datasets can contain multiple correlated images from the same lesion or patient. Samples were grouped by patient-level identifier when available, or otherwise by lesion identifier, ensuring that no group was shared between the training and test subsets. This preserves evaluation independence and reduces the risk of overly optimistic performance estimates.

\subsection{Data Preprocessing and Representation}

This subsection describes the preprocessing and feature representation steps applied to both dermoscopic images and structured clinical metadata prior to model training. The objective is to standardize inputs, reduce noise introduced by acquisition variability, and convert heterogeneous metadata fields into numerical representations suitable for multimodal learning.

\subsubsection{Image Preprocessing}

All dermoscopic images are converted into a consistent input format and resized to a fixed spatial resolution of $224 \times 224$ pixels. During training, data augmentation is applied to improve invariance to common visual variations, including random horizontal flipping, small-angle rotation, and color jittering. Random erasing is further used as a regularization technique to reduce reliance on localized spurious cues. For validation, deterministic resizing and normalization are applied without augmentation. Images are normalized using ImageNet statistics to align the input distribution with the pretrained vision backbone.

\subsubsection{Clinical Metadata Encoding}

Structured clinical attributes are transformed into a compact numerical feature vector for multimodal fusion. Age values are imputed for missing entries using the median of the training set and scaled to the $[0,1]$ range. Categorical variables, including sex and anatomical site, are encoded using one-hot representations, with an additional category reserved for unknown or missing values. The final metadata vector is formed by concatenating the normalized age feature with the one-hot encoded categorical features, producing a fixed-length representation for each sample.

\subsection{Multimodal Transformer Architecture}

The proposed model is a dual-branch multimodal network that integrates dermoscopic image features with structured clinical metadata, as illustrated in Fig.~\ref{multimodal_arch}. Given an input image $I \in \mathbb{R}^{H\times W\times 3}$ and a metadata vector $m \in \mathbb{R}^{d}$, the model learns modality-specific representations and combines them through feature-level fusion to obtain a final class probability distribution over $C$ lesion categories.

\subsubsection{Image Feature Extractor}
The visual branch utilizes a Swin Transformer backbone as the image feature extractor. Let $E_{\text{img}}(\cdot)$ denote the image encoder. The backbone maps the dermoscopic image to a compact embedding, as given in Eq.~\eqref{eq:z_img}.
\begin{equation}
\mathbf{z}_{\text{img}} = E_{\text{img}}(I).
\label{eq:z_img}
\end{equation}
where $I \in \mathbb{R}^{H\times W\times 3}$ denotes the input dermoscopic image and $\mathbf{z}_{\text{img}} \in \mathbb{R}^{d_{\text{img}}}$ denotes the resulting image feature embedding of dimensionality $d_{\text{img}}$. The classification layer of the backbone is removed, and the extracted representation is treated as a global descriptor that captures lesion appearance through hierarchical self-attention. The image encoder is fine-tuned simultaneously with the remaining components of the proposed multimodal model.

\subsubsection{Metadata Feature Encoder}
Clinical metadata is transformed into a learnable feature representation through a multilayer perceptron, denoted by $E_{\text{meta}}(\cdot)$. The metadata embedding is computed as shown in Eq.~\eqref{eq:z_meta}.
\begin{equation}
\mathbf{z}_{\text{meta}} = E_{\text{meta}}(m).
\label{eq:z_meta}
\end{equation}
where $m \in \mathbb{R}^{d}$ represents the encoded clinical metadata vector and $\mathbf{z}_{\text{meta}} \in \mathbb{R}^{d_{\text{meta}}}$ is the metadata feature embedding with dimensionality $d_{\text{meta}}$. The encoder is implemented using fully connected layers with non-linear activation, normalization, and dropout regularization to produce a compact representation of patient context.

\subsubsection{Fusion and Classification Head}
The modality-specific embeddings from Eq.~\eqref{eq:z_img} and Eq.~\eqref{eq:z_meta} are integrated through feature concatenation, as defined in Eq.~\eqref{eq:fusion}.
\begin{equation}
\mathbf{z} = [\mathbf{z}_{\text{img}} \, \| \, \mathbf{z}_{\text{meta}}].
\label{eq:fusion}
\end{equation}
where $[\cdot \| \cdot]$ denotes vector concatenation and $\mathbf{z} \in \mathbb{R}^{d_{\text{img}} + d_{\text{meta}}}$ is the fused representation. 

Then the fused representation is passed to a classification head $g(\cdot)$ to produce class logits, given by $\mathbf{s} = g(\mathbf{z})$, where $\mathbf{s} \in \mathbb{R}^{C}$. The logits are then transformed into a probability distribution over the $C$ lesion categories using the softmax function, $\hat{\mathbf{p}} = \mathrm{softmax}(\mathbf{s})$, where $\hat{\mathbf{p}} \in [0,1]^C$ denotes the predicted class probability vector and $\hat{p}_c$ represents the predicted probability of class $c$. This fusion design integrates complementary visual and clinical information while maintaining a lightweight classification head suitable for settings with limited metadata availability.


\subsection{Imbalance Mitigation Strategy}

\begin{table}[t]
\centering
\caption{Training data augmentation techniques and corresponding parameter settings.}
\label{aug_settings}
\renewcommand{\arraystretch}{1.1}
\setlength{\tabcolsep}{4pt}
\begin{tabular}{p{2cm}p{6cm}}
\hline
\textbf{Augmentation} & \textbf{Setting} \\
\hline
Resize-crop & Resize to $256\times256$ and randomly crop to $224\times224$ \\
Horizontal flip & Flip probability $=0.5$ \\
Rotation & Maximum rotation angle $=\pm 15^{\circ}$ \\
Color jitter & Brightness $=0.15$, Contrast $=0.15$ \newline Saturation $=0.15$, Hue $=0.05$ \\
Random erasing & Erasing probability $=0.25$, Scale $=0.02$ - $0.12$ \newline Aspect ratio $=0.3$ - $3.3$, Fill value: random \\
\hline
\end{tabular}
\end{table}

Class imbalance in the dataset is addressed through an integrated sampling, loss-weighting, and augmentation strategy during training. The data loader increases the representation of underrepresented classes within mini-batches, improving minority-class exposure across training iterations. Class-weighted cross-entropy is used to set higher penalties on minority-class misclassifications and reduce majority-class dominance during optimization. In addition, training-time augmentation, including resize-crop preprocessing, spatial and photometric transformations, random erasing, and regularization, is applied to improve generalization and limit overfitting to dominant visual patterns. Resize and normalization are applied consistently across all inputs, whereas stochastic augmentations are restricted to the training set, and validation preprocessing is kept deterministic. The complete configuration is summarized in Table~\ref{aug_settings}.

\subsection{Optimization and Training Procedure}

Model training was performed using an adaptive gradient-based optimizer with decoupled weight decay and a cosine learning-rate schedule to provide smooth learning-rate decay throughout training. Mixed-precision computation was utilized to improve graphics processing unit (GPU) efficiency, while gradient norm clipping was applied to stabilize parameter updates. Model selection was based on validation macro-F1 score, with early stopping used when validation performance failed to improve for several consecutive epochs. The experiments were conducted in Google Colab using an NVIDIA T4 GPU, and the proposed PyTorch-based multimodal framework was trained for approximately 95 epochs, requiring nearly 4.5 hours of total training time.



\section{Experimental Results and Evaluation}
\label{sec4}

This section reports quantitative performance results and analyzes model reliability through calibration, uncertainty estimation, and interpretability analysis to facilitate transparent clinical decision-making.

\subsection{Overall Classification Performance}

\begin{figure}[b]
    \centering
    \includegraphics[width=0.6\linewidth]{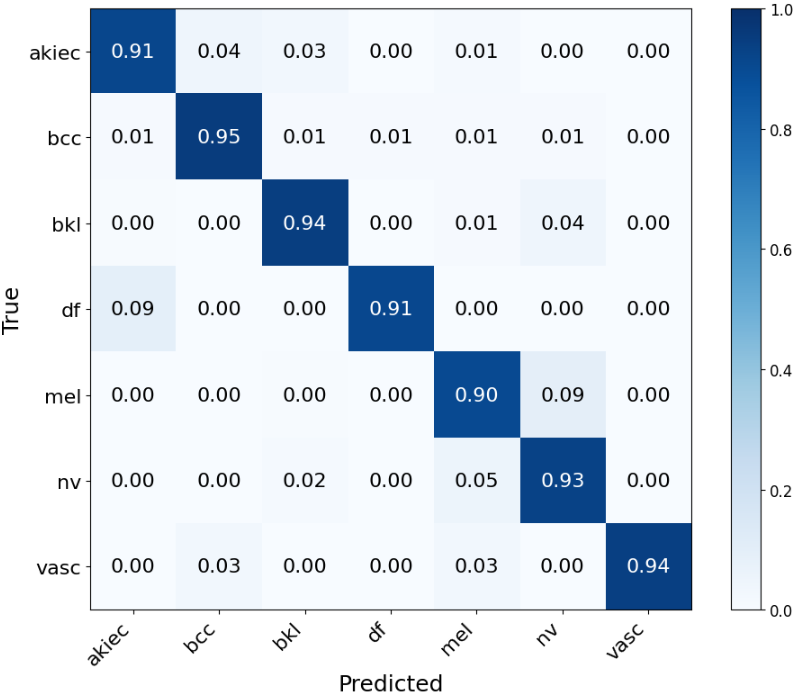}
    \caption{Row-normalized confusion matrix for the proposed classifier on the test set.}
    \label{confusion_matrix}
\end{figure}

\begin{table}[t]
\centering
\caption{Per-class classification and calibration results on the test set after temperature scaling ($T=2.5025$), including negative log-likelihood (NLL), one-vs-rest (OvR) Brier score, and expected calibration error (ECE).}
\label{cls_report}
\setlength{\tabcolsep}{5pt}
\renewcommand{\arraystretch}{1.05}
\begin{tabular}{lrrrr}
\hline
\multicolumn{5}{c}{\textbf{Classification metrics}} \\
\hline
\textbf{Class} & \textbf{Precision} & \textbf{Recall} & \textbf{F1-score} & \textbf{Support} \\
\hline
akiec & 0.9118 & 0.9118 & 0.9118 & 68  \\
bcc   & 0.9135 & 0.9500 & 0.9314 & 100 \\
bkl   & 0.8853 & 0.9415 & 0.9125 & 205 \\
df    & 0.9524 & 0.9091 & 0.9302 & 22  \\
mel   & 0.7220 & 0.9009 & 0.8016 & 222 \\
nv    & 0.9760 & 0.9259 & 0.9503 & 1363 \\
vasc  & 0.9697 & 0.9412 & 0.9552 & 34  \\
\hline
\textbf{Macro avg} & 0.9044 & 0.9258 & 0.9133 & 2014 \\
\textbf{Accuracy}  & \multicolumn{4}{r}{\textbf{0.9255}}\\
\hline
\multicolumn{5}{c}{\textbf{Calibration metrics (after temperature scaling)}} \\
\hline
\textbf{Class} & \textbf{NLL} & \textbf{Brier (OvR)} & \textbf{ECE (OvR)} & \textbf{Avg Conf.} \\
\hline
akiec & 0.4493 & 0.004228 & 0.004599 & 0.9529 \\
bcc   & 0.1770 & 0.005552 & 0.004231 & 0.9591 \\
bkl   & 0.2285 & 0.014810 & 0.013760 & 0.9199 \\
df    & 0.7350 & 0.001749 & 0.002161 & 0.9290 \\
mel   & 0.2570 & 0.038228 & 0.044290 & 0.8382 \\
nv    & 0.1996 & 0.051780 & 0.057121 & 0.9322 \\
vasc  & 0.2420 & 0.001042 & 0.001559 & 0.9454 \\
\hline
\end{tabular}
\end{table}

The overall classification performance of the proposed multimodal model is summarized in Table~\ref{cls_report}, while the corresponding confusion patterns are illustrated in Fig.~\ref{confusion_matrix}. The row-normalized confusion matrix indicates strong diagonal dominance, indicating consistent class-wise recognition and limited confusion across categories. Consistent with this observation, the model achieves an overall test accuracy of 0.9255 ($\approx 93\%$) and a macro-averaged F1-score of 0.9133, demonstrating strong performance across imbalanced classes. Per-class precision and recall values further confirm that the learned decision boundaries are effective across lesion types under the group-aware evaluation scheme. A comparison with related studies conducted on the same HAM10000 dataset is presented in Table~\ref{tab:comparison}, where the proposed approach shows competitive performance and improved practical reliability for medical artificial intelligence (AI) applications.

\begin{table}[b]
\centering
\caption{Comparison of the proposed method with related studies on the HAM10000 dataset.}
\label{tab:comparison}
\renewcommand{\arraystretch}{1.15}
\setlength{\tabcolsep}{5pt}
\begin{tabular}{c p{2.2cm} p{3.7cm} p{8mm}}
\hline
\textbf{Ref.} & \textbf{Input Modality} & \centering\textbf{Model} & \textbf{Acc.(\%)} \\
\hline
\cite{dubey2023unet} & Image only & Two-stage U-Net-SVM & 78.0 \\
\hline
\cite{salian2020skin} & Image only & MobileNet, VGG-16 & 80.61 \\
\hline
\cite{ameri2020deep} & Image only & AlexNet & 84.0 \\
\hline
\cite{srinivasu2021classification} & Image only & MobileNet + long short-term memory (LSTM) & 85.0 \\
\hline
\multirow{4}{*}{\cite{rahman2020transfer}} 
& \multirow{4}{*}{Image only} & ResNet & 78.0 \\
\cline{3-4}
& & Xception & 82.0 \\
\cline{3-4}
& & DenseNet & 82.0 \\
\cline{3-4}
& & Weighted ensemble & 85.8 \\
\hline
\cite{yao2021single} & Image only & RegNetY-3.2GF & 85.8 \\
\hline
\cite{khan2024integration} & Image + metadata & CNN-LSTM multimodal fusion & 92.0 \\
\hline
-- & \textbf{Image + metadata} & \textbf{Proposed model} & \textbf{92.55} \\
\hline
\end{tabular}
\end{table}



\subsection{Impact of Class Imbalance Handling}

\begin{figure*}[t]
    \centering
    \includegraphics[width=0.85\textwidth]{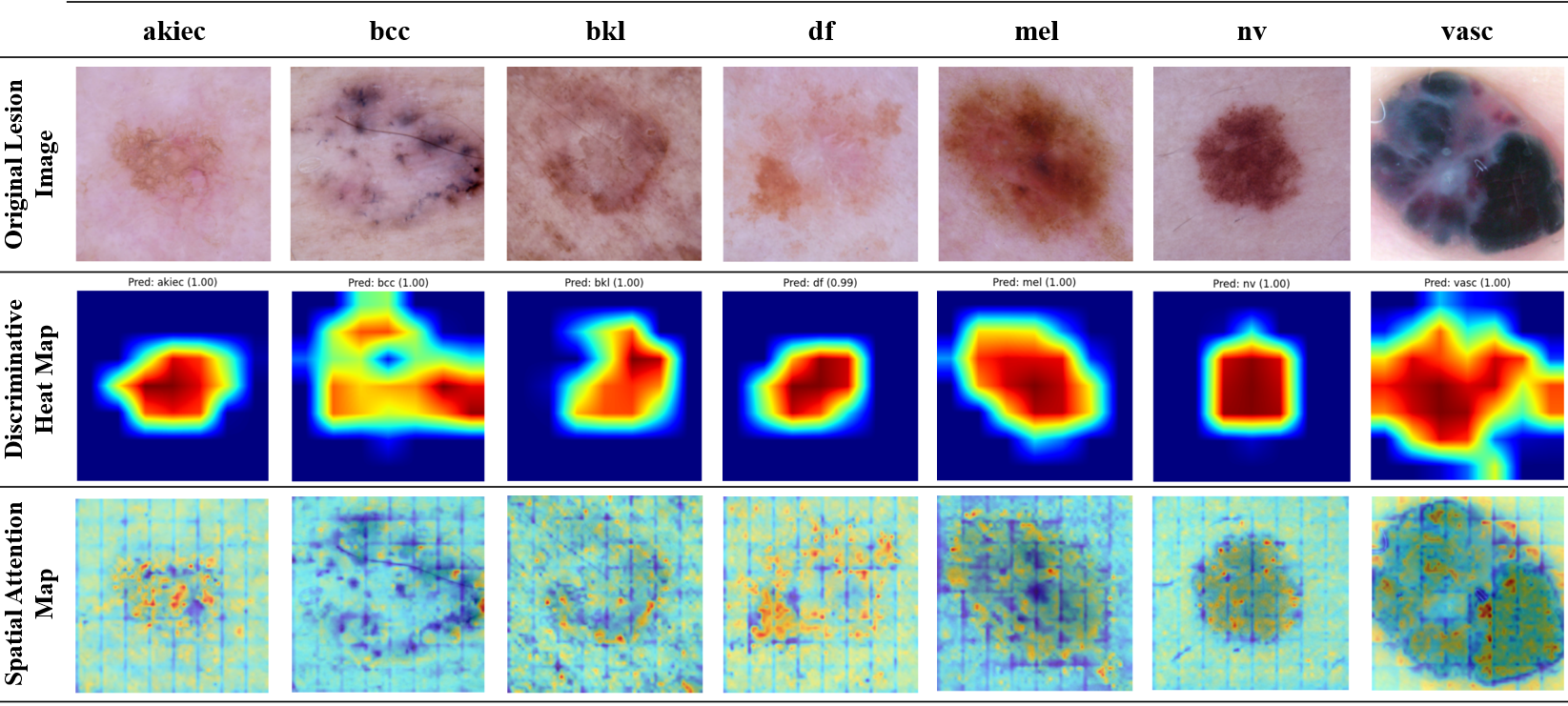}
    \caption{Explainable AI visualizations for the proposed multimodal transformer classifier, showing the dermoscopic input, a class-discriminative attribution heatmap of regions most influential to the predicted class and confidence, and a patch-level transformer attention or feature-focus map indicating how spatial evidence is integrated during inference.}
    \label{qualitative_interpretability}
\end{figure*}

\begin{figure}[b]
    \centering
    \includegraphics[width=1\linewidth]{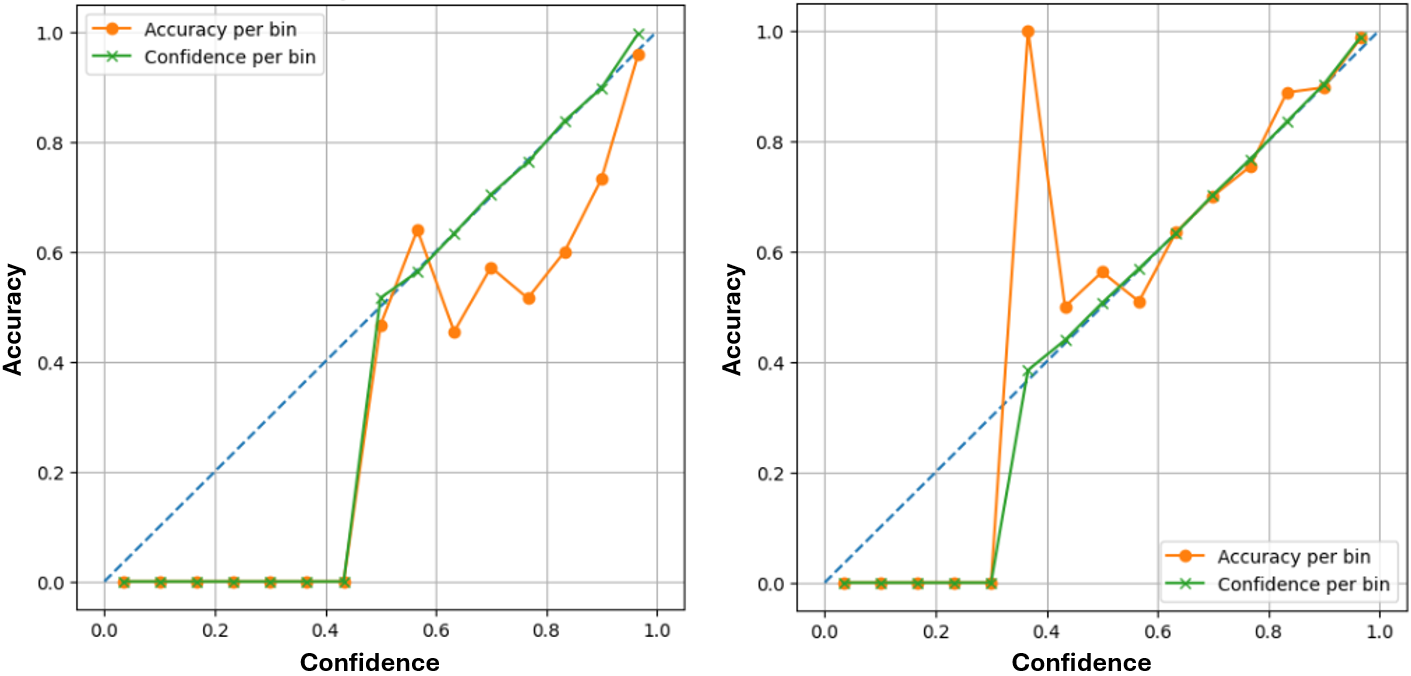}
    \caption{Reliability diagrams comparing uncalibrated predictions (left) and temperature-scaled calibrated predictions with $T=2.5025$ (right), showing the relationship between mean confidence and observed accuracy across confidence bins relative to the ideal identity line.}
    \label{calibrated_matrix}
\end{figure}

Given the substantial class imbalance shown in Table~\ref{ham10000_class_dist}, class imbalance mitigation was incorporated into the training pipeline to reduce the influence of the majority class and to improve the optimization of underrepresented categories. Under this setting, the model demonstrates strong minority class recognition on the test set, achieving recall values of 0.9091 for df, 0.9412 for vasc, and 0.9118 for akiec with 22, 34, and 68 test samples, respectively, with corresponding F1 scores of 0.9302, 0.9552, and 0.9118 as summarized in Table~\ref{cls_report}. In addition, bcc attains a recall of 0.9500 with 100 test samples and an F1 score of 0.9314, indicating strong discrimination for malignant categories under imbalanced class frequencies. Overall, these results are consistent with the intended effect of class-aware rebalancing through weighted sampling and class-weighted optimization, which promotes more uniform learning across lesion categories.

\subsection{Comparison with Unimodal Baselines}

The comparison in Table~\ref{unimodal_comparison} shows that the proposed multimodal framework outperforms both unimodal baselines. The image-only model achieved strong performance, while the metadata-only model showed limited classification ability. These results indicate that dermoscopic images provide the main discriminative information and that the addition of structured clinical metadata further improves overall performance.

\begin{table}[t]
\centering
\caption{Comparison of unimodal baselines and the proposed multimodal framework.}
\label{unimodal_comparison}
\begin{tabular}{lcrr}
\toprule
\textbf{Model} & \textbf{Input} & \textbf{Accuracy} & \textbf{F1-score} \\
\midrule
Metadata only & Metadata & 0.1113 & 0.1016 \\
Image only & Image & 0.8862 & 0.8643 \\
\textbf{Proposed multimodal} & \textbf{Image + metadata} & \textbf{0.9255} & \textbf{0.9133} \\
\bottomrule
\end{tabular}
\end{table}

\subsection{Qualitative Explainability Assessment}

We conduct a qualitative XAI analysis using class discriminative attribution maps and transformer-derived spatial focus visualizations to support model transparency and clinical plausibility, as illustrated in Fig.~\ref{qualitative_interpretability}. The attribution heatmaps localize regions that contribute most strongly to the predicted class and associated confidence, indicating that the model primarily uses lesion-specific structures instead of background artifacts such as hair, surrounding skin tone, or acquisition-related noise.

In addition, patch-level attention or feature focus maps provide complementary evidence of how the transformer aggregates localized cues into a global decision during inference. Overall, these qualitative findings are consistent with the quantitative results in Fig.~\ref{confusion_matrix} and offer an interpretable perspective on the model decision process, particularly for categories with overlapping visual characteristics.

\subsection{Model Calibration Analysis and Predictive Uncertainty}

We used temperature scaling to improve the probabilistic calibration of our proposed classifier, and its impact was assessed using reliability diagrams with 15 equal-width confidence bins spanning ([0,1]). As illustrated in Fig.~\ref{calibrated_matrix}, for each bin the mean predicted confidence (maximum predicted probability) is compared against the empirical accuracy (fraction of correct predictions), with the identity line indicating perfect calibration. Before calibration, the model demonstrates a noticeable confidence–accuracy discrepancy, with an overall uncalibrated ECE of 0.0518 (5.18\%), indicating a non-trivial average mismatch between predicted confidence and observed correctness under the bin-weighted ECE definition. After temperature scaling, the calibrated reliability curve closely tracks the identity line across most confidence ranges, and the overall ECE decreases to 0.0091 (0.91\%), confirming a substantial improvement in the trustworthiness and interpretability of the predicted probabilities.

Class-wise calibration performance after temperature scaling is reported in Table~\ref{cls_report} using NLL, Brier score (OvR), and ECE (OvR). The results show low calibration error for multiple classes, accompanied by low NLL values, indicating that the calibrated model assigns high and reliable probabilities to correct predictions. Across classes, the model maintains a high average confidence of approximately 0.84-0.96, while the per-class OvR metrics provide a fine-grained characterization of calibration quality and predictive certainty for each lesion category. Overall, the combined reliability analysis (Fig.~\ref{calibrated_matrix}) and per-class calibration metrics (Table~\ref{cls_report}) demonstrate that temperature scaling produces well-aligned confidence estimates and improves the interpretability of probability outputs for downstream clinical decision support.

\section{Conclusion and Future Work}

This study shows that multimodal learning offers a robust and reliable approach to automated skin lesion classification by integrating dermoscopic image features with structured clinical metadata. The proposed framework achieved strong performance on the HAM10000 dataset, with a test accuracy of 92.55\% and a macro F1-score of 91.33\%, while also maintaining good performance on minority classes. In addition to classification performance, the study emphasizes reliability in medical artificial intelligence, as temperature scaling significantly improved calibration by reducing the ECE from 5.18\% to 0.91\%, uncertainty estimation improved confidence assessment, and XAI analysis confirmed that the model consistently attended to clinically relevant lesion regions. Future work will focus on external validation, the inclusion of richer clinical attributes, and the development of uncertainty-aware referral mechanisms to further strengthen practical clinical applicability.

\bibliographystyle{IEEEtran}
\bibliography{references}

@article{inthiyaz2023skin,
  title={Skin disease detection using deep learning},
  author={Inthiyaz, Syed and Altahan, Baraa Riyadh and Ahammad, Sk Hasane and Rajesh, V and Kalangi, Ruth Ramya and Smirani, Lassaad K and Hossain, Md Amzad and Rashed, Ahmed Nabih Zaki},
  journal={Advances in Engineering Software},
  volume={175},
  pages={103361},
  year={2023},
  publisher={Elsevier}
}

@inproceedings{demir2019early,
  title={Early detection of skin cancer using deep learning architectures: resnet-101 and inception-v3},
  author={Demir, Ahmet and Yilmaz, Feyza and Kose, Onur},
  booktitle={2019 medical technologies congress (TIPTEKNO)},
  pages={1--4},
  year={2019},
  organization={IEEE}
}

@article{bechelli2022machine,
  title={Machine learning and deep learning algorithms for skin cancer classification from dermoscopic images},
  author={Bechelli, Solene and Delhommelle, Jerome},
  journal={Bioengineering},
  volume={9},
  number={3},
  pages={97},
  year={2022},
  publisher={MDPI}
}

@article{gajera2023comprehensive,
  title={A comprehensive analysis of dermoscopy images for melanoma detection via deep CNN features},
  author={Gajera, Himanshu K and Nayak, Deepak Ranjan and Zaveri, Mukesh A},
  journal={Biomedical Signal Processing and Control},
  volume={79},
  pages={104186},
  year={2023},
  publisher={Elsevier}
}

@article{shinde2022squeeze,
  title={Squeeze-MNet: precise skin cancer detection model for low computing IoT devices using transfer learning},
  author={Shinde, Rupali Kiran and Alam, Md Shahinur and Hossain, Md Biddut and Md Imtiaz, Shariar and Kim, JoonHyun and Padwal, Anuja Anil and Kim, Nam},
  journal={Cancers},
  volume={15},
  number={1},
  pages={12},
  year={2022},
  publisher={MDPI}
}

@article{rashid2022skin,
  title={Skin cancer disease detection using transfer learning technique},
  author={Rashid, Javed and Ishfaq, Maryam and Ali, Ghulam and Saeed, Muhammad R and Hussain, Mubasher and Alkhalifah, Tamim and Alturise, Fahad and Samand, Noor},
  journal={Applied Sciences},
  volume={12},
  number={11},
  pages={5714},
  year={2022},
  publisher={MDPI}
}

@article{alenezi2023wavelet,
  title={Wavelet transform based deep residual neural network and ReLU based Extreme Learning Machine for skin lesion classification},
  author={Alenezi, Fayadh and Armghan, Ammar and Polat, Kemal},
  journal={Expert Systems with Applications},
  volume={213},
  pages={119064},
  year={2023},
  publisher={Elsevier}
}

@article{alenezi2023multi,
  title={A multi-stage melanoma recognition framework with deep residual neural network and hyperparameter optimization-based decision support in dermoscopy images},
  author={Alenezi, Fayadh and Armghan, Ammar and Polat, Kemal},
  journal={Expert Systems with Applications},
  volume={215},
  pages={119352},
  year={2023},
  publisher={Elsevier}
}

@article{bassel2022automatic,
  title={Automatic malignant and benign skin cancer classification using a hybrid deep learning approach},
  author={Bassel, Atheer and Abdulkareem, Amjed Basil and Alyasseri, Zaid Abdi Alkareem and Sani, Nor Samsiah and Mohammed, Husam Jasim},
  journal={Diagnostics},
  volume={12},
  number={10},
  pages={2472},
  year={2022},
  publisher={MDPI}
}

@article{reis2022insinet,
  title={InSiNet: a deep convolutional approach to skin cancer detection and segmentation},
  author={Reis, Hatice Catal and Turk, Veysel and Khoshelham, Kourosh and Kaya, Serhat},
  journal={Medical \& Biological Engineering \& Computing},
  volume={60},
  number={3},
  pages={643--662},
  year={2022},
  publisher={Springer}
}

@article{shorfuzzaman2022explainable,
  title={An explainable stacked ensemble of deep learning models for improved melanoma skin cancer detection},
  author={Shorfuzzaman, Mohammad},
  journal={Multimedia Systems},
  volume={28},
  number={4},
  pages={1309--1323},
  year={2022},
  publisher={Springer}
}

@article{bian2022skin,
  title={Skin lesion image classification method based on extension theory and deep learning},
  author={Bian, Xiaofei and Pan, Haiwei and Zhang, Kejia and Li, Pengyuan and Li, Jinbao and Chen, Chunling},
  journal={Multimedia Tools and Applications},
  volume={81},
  number={12},
  pages={16389--16409},
  year={2022},
  publisher={Springer}
}

@article{ahmed2020malignant,
  title={Malignant melanoma: skin cancer- diagnosis, prevention, and treatment},
  author={Ahmed, Bilal and Qadir, Muhammad Imran and Ghafoor, Saba},
  journal={Critical reviews™ in eukaryotic gene expression},
  volume={30},
  number={4},
  year={2020},
  publisher={Begel House Inc.}
}

@article{wen2022characteristics,
  title={Characteristics of publicly available skin cancer image datasets: a systematic review},
  author={Wen, David and Khan, Saad M and Xu, Antonio Ji and Ibrahim, Hussein and Smith, Luke and Caballero, Jose and Zepeda, Luis and de Blas Perez, Carlos and Denniston, Alastair K and Liu, Xiaoxuan and others},
  journal={The Lancet Digital Health},
  volume={4},
  number={1},
  pages={e64--e74},
  year={2022},
  publisher={Elsevier}
}

@inproceedings{rodrigo2026benchmarking,
  title={Benchmarking CNN and Transformer-Based Object Detectors for UAV Solar Panel Inspection},
  author={Rodrigo, Ashen and Munasinghe, Isuru and Sanjeewani, Pubudu and Perera, Asanka},
  booktitle={2026 International Conference on Image Processing and Robotics (ICIPRoB)},
  pages={1--6},
  year={2026},
  organization={IEEE}
}

@article{yao2021single,
  title={Single model deep learning on imbalanced small datasets for skin lesion classification},
  author={Yao, Peng and Shen, Shuwei and Xu, Mengjuan and Liu, Peng and Zhang, Fan and Xing, Jinyu and Shao, Pengfei and Kaffenberger, Benjamin and Xu, Ronald X},
  journal={IEEE transactions on medical imaging},
  volume={41},
  number={5},
  pages={1242--1254},
  year={2021},
  publisher={IEEE}
}

@article{tschandl2018ham10000,
  title={The HAM10000 dataset, a large collection of multi-source dermatoscopic images of common pigmented skin lesions},
  author={Tschandl, Philipp and Rosendahl, Cliff and Kittler, Harald},
  journal={Scientific data},
  volume={5},
  number={1},
  pages={180161},
  year={2018},
  publisher={Nature Publishing Group}
}

@article{ameri2020deep,
  title={A deep learning approach to skin cancer detection in dermoscopy images},
  author={Ameri, A},
  journal={Journal of biomedical physics \& engineering},
  volume={10},
  number={6},
  pages={801},
  year={2020}
}

@inproceedings{salian2020skin,
  title={Skin lesion classification using deep learning architectures},
  author={Salian, Abhishek C and Vaze, Shalaka and Singh, Pragya and Shaikh, Gulam Nasir and Chapaneri, Santosh and Jayaswal, Deepak},
  booktitle={2020 3rd International conference on communication system, computing and IT applications (CSCITA)},
  pages={168--173},
  year={2020},
  organization={IEEE}
}

@inproceedings{rahman2020transfer,
  title={A transfer learning based approach for skin lesion classification from imbalanced data},
  author={Rahman, Zillur and Ami, Amit Mazumder},
  booktitle={2020 11th international conference on electrical and computer engineering (ICECE)},
  pages={65--68},
  year={2020},
  organization={IEEE}
}

@article{srinivasu2021classification,
  title={Classification of skin disease using deep learning neural networks with MobileNet V2 and LSTM},
  author={Srinivasu, Parvathaneni Naga and SivaSai, Jalluri Gnana and Ijaz, Muhammad Fazal and Bhoi, Akash Kumar and Kim, Wonjoon and Kang, James Jin},
  journal={Sensors},
  volume={21},
  number={8},
  pages={2852},
  year={2021},
  publisher={MDPI}
}

@article{khan2024integration,
  title={Integration of Multimodal Data Sources for Enhanced Skin Disease Classification and Cancer Prediction: A Study Leveraging Pre-Trained Models on HAM\_10000 Metadata and Squamous Cell Carcinoma (SCC) Images.},
  author={Khan, Aamir and Chandran, Nisha and Gangodkar, DR},
  year={2024}
}

@inproceedings{dubey2023unet,
  title={UNet Segmentation based Effective Skin Lesion Detection using Deep Learning},
  author={Dubey, Arun Kumar and Jain, Achin and Panwar, Arvind and Kumar, Manish and Taneja, Harsh and Lamba, Puneet Singh},
  booktitle={2023 International Conference on Communication, Security and Artificial Intelligence (ICCSAI)},
  pages={470--474},
  year={2023},
  organization={IEEE}
}

@article{rodrigo2025vision,
  title={Vision-Based Object Detection for UAV Solar Panel Inspection Using an Enhanced Defects Dataset},
  author={Rodrigo, Ashen and Munasinghe, Isuru and Perera, Asanka},
  journal={arXiv e-prints},
  pages={arXiv--2509},
  year={2025}
}

\end{document}